\documentclass{article} 
\usepackage{iclr2027_conference,times}

\usepackage{amsmath,amsfonts,bm}

\def\eqref#1{equation~\ref{#1}}

\def\1{\bm{1}}

\DeclareMathAlphabet{\mathsfit}{\encodingdefault}{\sfdefault}{m}{sl}
\SetMathAlphabet{\mathsfit}{bold}{\encodingdefault}{\sfdefault}{bx}{n}

\usepackage{graphicx}
\usepackage{hyperref}
\usepackage{url}
\usepackage{mathrsfs}
\usepackage{booktabs}
\usepackage[table,xcdraw]{xcolor}
\usepackage{wrapfig}

\usepackage{etoolbox}

\iclrfinalcopy

\title{Uncovering Uncontrolled Repetition through Residual Stream Dynamics}

\author{
 \textbf{Yuanhe Zhang\textsuperscript{1}}, 
 \textbf{Xinyao Zhou\textsuperscript{1}}, 
 \textbf{Haoran Gao\textsuperscript{2}}, 
 \textbf{Yuyao Zhang\textsuperscript{2}}, 
 \\
 \textbf{Zhenhong Zhou\textsuperscript{3}}, 
 \textbf{Fanyu Meng\textsuperscript{2}},
 \textbf{Li Sun\textsuperscript{1}},
 \textbf{Sen Su\textsuperscript{1, 4,  $^\dagger$}} 
\\ \textsuperscript{\rm 1}Beijing University of Posts and Telecommunications
\\ \textsuperscript{\rm 2}JIUTIAN Research
\\ \textsuperscript{\rm 3}Nanyang Technological University
\\ \textsuperscript{\rm 4}Chongqing University of Posts and Telecommunications
\\ \{charmes-zhang, susen\}@bupt.edu.cn;
}

\begin{document}

\maketitle
\begingroup
\renewcommand\thefootnote{}\footnotemark
\footnotetext{$\dagger$ indicates corresponding author.}
\endgroup

\begin{abstract}

Uncontrolled repetition can prolong autoregressive generation in large language models (LLMs) and enable resource consumption attacks.
Prior analyses of repetitive generation have identified strongly activated features in intermediate and late layers.
However, how uncontrolled repetition activity emerges and develops before becoming prominent in these layers remains insufficiently understood.
In this paper, we investigate this question primarily in large vision-language models (LVLMs), which support a richer set of uncontrolled repetitions through both visual and textual inputs. 
We propose \textit{Tokenwise Residual Comparison} (TRC), a method that identifies and localizes anomalies associated with repetition from residual dynamics during generation.
TRC compares attention and multilayer perceptron writes to the residual stream across generated tokens to identify patterns associated with repetition. 
It then selectively suppresses coordinates in the residual stream at the identified layer.
Experiments show that TRC effectively mitigates uncontrolled repetition, reducing loop rates by 57\% on average.
Our analysis further shows that repetition semantics emerge in shallow layers and propagate through the residual stream, disrupting normal representations.
TRC also generalizes to large language models (LLMs) and large reasoning models (LRMs), where it consistently captures analogous repetition dynamics and achieves effective mitigation.
Our work broadens the study of repetitive generation from its prominent internal representations to earlier opportunities for intervention, providing insights for mitigating resource consumption attacks.
\end{abstract}

\section{Introduction}
\label{sec:introduction}

Language models can become trapped in repetitive generation, where recurring content disrupts the normal development of the response~\citep{hiraoka-inui-2025-repetition}.
Attackers can deliberately induce such repetition through textual or visual inputs to increase inference cost and latency, threatening service availability \citep{Li_2026,gao2025resourceconsumptionredteaminglarge}.
Prior studies have primarily analyzed repetition using manually constructed repetitive sequences \citep{yao-etal-2025-understanding,hiraoka-inui-2025-repetition}.
These analyses identify repetition-related features concentrated in intermediate and final layers \citep{hiraoka-inui-2025-repetition,yao-etal-2025-understanding}.
We argue that understanding these prominent representations also requires examining how uncontrolled repetition emerges across layers during model generation.
LVLMs provide a convenient setting for this analysis, as the continuous space of visual inputs facilitates the construction of examples that induce uncontrolled generation \citep{ICLR2024_4a6a5e2e,gao2025resourceconsumptionredteaminglarge}.
By varying visual and textual triggers, we can study a broader range of model-generated failure cases and examine repetition activity before it becomes prominent in deeper layers.

Existing mitigation strategies address repetition through decoding controls and internal activation interventions \citep{NEURIPS2022_148c0aee,yao-etal-2025-understanding, zhang2026resource}.
At the output level, nucleus sampling and repetition penalties counter degenerate continuations by changing token selection \citep{holtzman2020curiouscaseneuraltext,zhu-etal-2023-penalty}.
Output-budget constraints require balancing token cost against answer accuracy, while aggressive repetition penalties can impair legitimate generation \citep{han-etal-2025-token,gao2025resourceconsumptionredteaminglarge}.
At the activation level, prior studies identify repetition related neurons or features from constructed repetitive sequences and show that suppressing their activations can mitigate repetition \citep{hiraoka-inui-2025-repetition,yao-etal-2025-understanding}.
However, these approaches primarily identify representations after the repetitive semantics have become strongly activated, leading to their localization mainly in intermediate or later layers \citep{hiraoka-inui-2025-repetition,yao-etal-2025-understanding}.
Our question is whether early layers already exhibit signals of repetition that can support analysis.

In this paper, we investigate this question using repetition that naturally emerges from LVLM generation under visual and textual input triggers, rather than artificially constructed repetition in model outputs.
We propose \textit{Tokenwise Residual Comparison} (TRC), which characterizes uncontrolled repetition changes in residual stream contributions.
TRC compares residual stream contributions across generated sequence to derive tokenwise repetition signals at each layer.
It then identifies candidate layers based on the magnitude of these signals relative to benign references and their local consistency across layers.
TRC scores at the selected layer are then used to construct a mask that selectively suppresses anomalous residual stream contributions.
By comparing residual stream contributions along the generated sequence, TRC exposes fine grained changes associated with repetition and localizes them to specific layers and coordinates.

Experiments show that suppressing residual contributions at the layer with the minimum TRC score reduces repetition by 57\% on average. The same criterion consistently localizes repetition related changes to shallow layers, suggesting that the signals captured by TRC emerge well before repetition becomes prominent in later representations. TRC further generalizes to large language models (LLMs) and large reasoning models (LRMs), where it similarly mitigates repetitive generation and identifies corresponding changes at early layers. These findings indicate that fine grained residual changes in shallow layers provide effective targets for observing and suppressing uncontrolled repetition before its representations become dominant.

In summary, we introduce TRC, which uses features from residual connections to characterize tokenwise repetition signals across layers. TRC reveals that uncontrolled repetition already leaves distinguishable traces in shallow layers, where targeted suppression substantially mitigates repetitive generation under both visual and textual triggers in LVLMs. This behavior further generalizes to LLMs and LRMs, indicating that early repetition signals and their intervention effects are shared across different model families.

\begin{figure}[t]
\centering
\includegraphics[width=\textwidth]{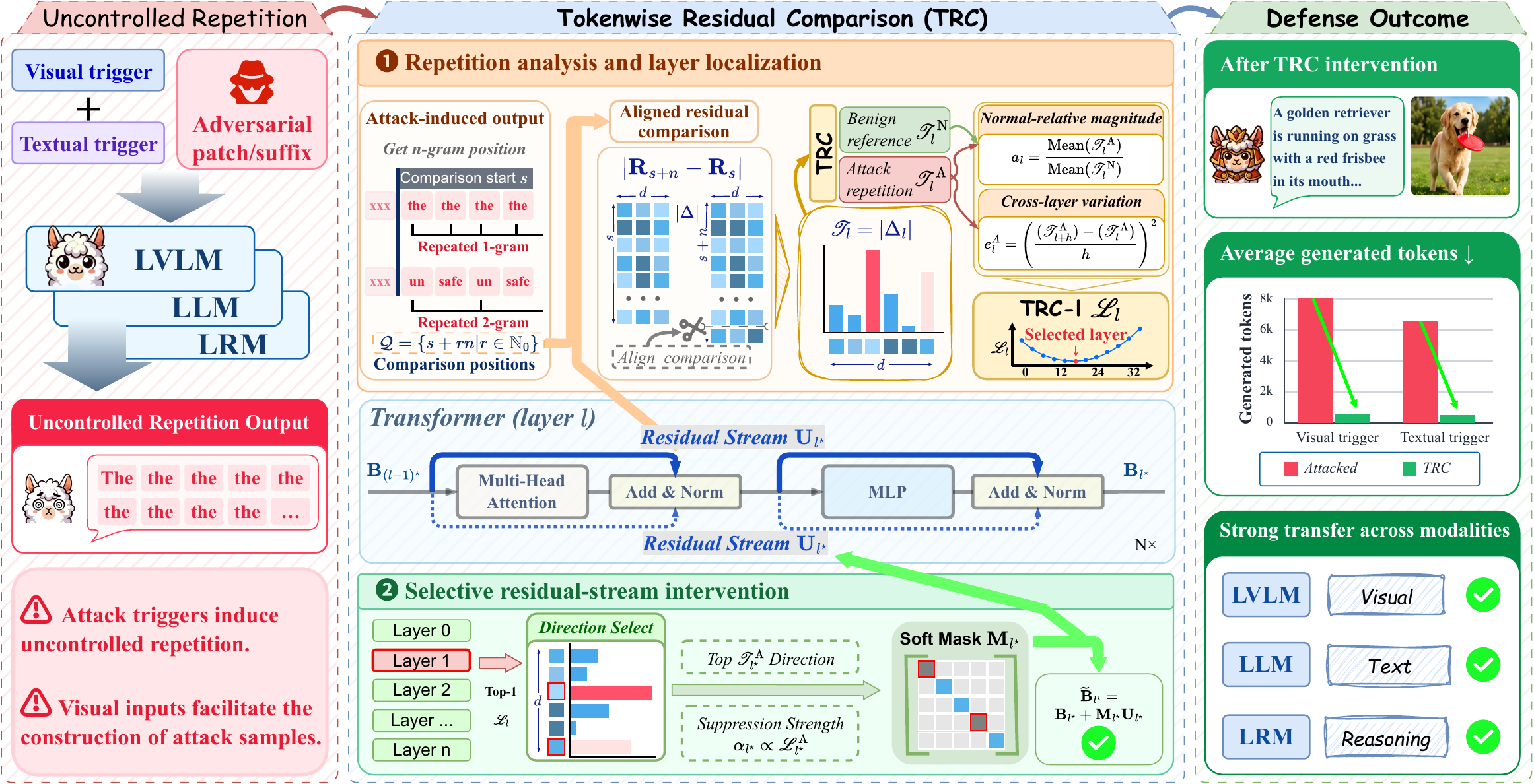}
\caption{Overview of \textbf{TRC}, showing uncontrolled repetition across model families (\textbf{\textit{Left}}), tokenwise residual comparison and selective intervention (\textbf{\textit{Middle}}), and mitigation outcomes(\textbf{\textit{Right}}).
}
\vspace{-9pt}
\label{fig:overview}
\end{figure}

\section{Related Work}
\label{sec:related_work}

\subsection{Input Modalities and Adversarial Example Construction}

Different input modalities provide distinct perturbation spaces for adversarial example construction. Text attacks operate over discrete tokens through efficient gradient guided modifications or direct contextual replacements, with continuous relaxations enabling optimization over discrete sequences\citep{ebrahimi-etal-2018-hotflip,li-etal-2020-bert-attack,garg-ramakrishnan-2020-bae,guo-etal-2021-gradient}. Visual and speech attacks instead optimize continuous pixels or waveforms under perceptual constraints \citep{Goodfellow2015,Carlini_2017,Carlini_2018,pmlr-v97-qin19a}. Multimodal attacks further exploit interactions across channels through coordinated perturbations and cross modal alignment \citep{zhang2022towards,NEURIPS2023_a5e3cf29,Lu_2023_ICCV}. In LVLMs, optimized pixels and typographic visual prompts can bypass textual safety alignment, while visual optimization can also prolong generation \citep{Qi_2024,Gong_2025,ICLR2024_4a6a5e2e}. This flexibility facilitates the construction of diverse visual and textual triggers for studying model generated repetition.

\subsection{Resource Consumption Attacks and Defenses}

Resource consumption attacks arise through different mechanisms across model types. Sponge examples increase energy use and latency, while DeepSloth and SlowBERT undermine early exit efficiency \citep{ilia2021spongeconference,Hong2021DeepSloth,zhang-etal-2023-slowbert}. In autoregressive models, optimized prompts, natural instructions, adversarial images, poisoned training data, and injected reasoning decoys can all prolong computation or induce excessive outputs \citep{ICLR2025_a815fe7c,chen-etal-2026-naturalsloth,ICLR2024_4a6a5e2e,gao2024denial,kumar2026overthinkslowdownattacksreasoning, zhang2025crabs}. Existing mitigation includes token budget control, filtering or paraphrasing external context, decoding strategies, and training objectives that discourage degeneration or repetition \citep{han-etal-2025-token,kumar2026overthinkslowdownattacksreasoning,holtzman2020curiouscaseneuraltext,zhu-etal-2023-penalty,NEURIPS2022_871cae8f,li-etal-2023-contrastive,welleck2019neural,li-etal-2020-dont,NEURIPS2022_148c0aee}. 
Existing internal analyses primarily capture prominent repetition activations, providing limited insight into when repetition first emerges during generation \citep{hiraoka-inui-2025-repetition,yao-etal-2025-understanding}. TRC instead tracks evolving semantic changes in residual stream contributions across the generated sequence to identify the emergence of repetition signals.
\section{Tokenwise Residual Comparison}
\label{sec:method}

Tokenwise Residual Comparison (TRC) identifies repetition signals by comparing differences in residual stream contributions, then selects residual coordinates for targeted suppression.
After specifying the calibration setting in Section~\ref{sec:method_setup}, we derive TRC scores and  layer localization scores (TRC-l) in Section~\ref{sec:method_metric}.
Section~\ref{sec:method_suppression} converts the selected TRC scores into a fixed soft mask applied at the selected layer in residual addition.

\subsection{Preliminaries and Problem Setup}
\label{sec:method_setup}

We consider an LVLM whose autoregressive generation can be driven into uncontrolled repetition.
The defender has access to the transformer backbone and records the contributions transmitted through its residual connections during generation.
Calibration uses a set of attack-induced repetition outputs $\mathcal{D}_{\mathrm{A}}$ and a separate set of benign reference outputs $\mathcal{D}_{\mathrm{N}}$.
Attention and multilayer perceptron (MLP) modules are analyzed independently within each model, so we omit the module index throughout.
The formulation also applies separately to text-only LLMs and LRMs.

Given an input request, the model generates one token at a time, conditioned on that request and its previously generated tokens.
We denote a completed output by $\mathbf{y}=(y_1,\ldots,y_T)$, where $y_t$ is the token at output position $t$ and $T\geq2$ is the output length.
The backbone has $L$ layers indexed by $l\in\{0,\ldots,L-1\}$, each with hidden dimension $d$.
At generation position $t$, $\mathbf{R}_{l,t}\in\mathbb{R}^{d\times S_t}$ denotes the recorded contribution of the selected module in residual addition, where $S_t$ is the total number of tokens from the input through generation position $t$.

Our goal is to characterize changes in the residual stream associated with uncontrolled repetition and use them to determine a localized intervention during generation. Given the recorded module contributions, we seek to identify layers and hidden coordinates that exhibit distinctive changes under repetition, and selectively suppress these contributions while preserving benign generation.

\subsection{TRC Scores and Layer Localization}
\label{sec:method_metric}

We measure changes in the contributions transmitted through residual connections across generation positions.
To identify the dominant repeated pattern, we search over $n$-grams with $n\in\{1,\ldots,\lfloor T/2\rfloor\}$.
For each $n$, we consider only $n$-grams occurring at least twice and measure the fraction of output tokens covered by their occurrences.
We then select the smallest $n$ whose most-covered repeated $n$-gram accounts for more than half of the output:
\begin{equation}
\begin{aligned}
n^\star=&\min\left\{n:\max_{g:\,|\mathcal{O}_n(g)|\ge2}\frac{\left|\bigcup_{t\in\mathcal{O}_n(g)}\{t,\ldots,t+n-1\}\right|}{T}>0.5\right\},\\
g^\star=&\operatorname{arg\,max}_{g:\,|\mathcal{O}_{n^\star}(g)|\ge2}\left|\bigcup_{t\in\mathcal{O}_{n^\star}(g)}\{t,\ldots,t+n^\star-1\}\right|,
\qquad
\mathcal{P}^\star=\mathcal{O}_{n^\star}(g^\star),
\end{aligned}
\label{eq:repeated_ngram}
\end{equation}
where $g=(g_1,\ldots,g_n)$ denotes an $n$-gram in $\mathbf{y}$, and
$\mathcal{O}_n(g)=\{t\in\{1,\ldots,T-n+1\}:(y_t,\ldots,y_{t+n-1})=g\}$
denotes the set of starting positions at which $g$ occurs in the output.
Thus, $n^\star$ gives the minimum $n$ for which a repeated $n$-gram covers more than $50\%$ of the output, while $\mathcal{P}^\star$ records all starting positions of the selected repeated $n$-gram.
We then determine the comparison start $s$, token spacing $q$, and comparison positions $\mathcal{Q}$ as:
\begin{equation}
(s,q)=
\begin{cases}
(\min\mathcal{P}^\star,n^\star), & \text{if } n^\star \text{ is defined},\\
(1,1), & \text{otherwise},
\end{cases}
\qquad
\mathcal{Q}=\{s+rq|r\in\mathbb{N}_0,\ s+(r+1)q\le T\}.
\label{eq:comparison_setup}
\end{equation}
The set $\mathbb{N}_0$ contains the nonnegative integers, and each $t\in\mathcal{Q}$ serves as the earlier position in a pairwise comparison with $(t,t+q)$.

Since $S_{t+q}=S_t+q$, we align each comparison pair by retaining the trailing columns corresponding to the newly generated tokens in $\mathbf{R}{l,t}$ and $\mathbf{R}{l,t+q}$. We then compute the element-wise absolute difference between these aligned module contributions:
\begin{equation}
\mathbf{v}_l
=
\frac{1}{|\mathcal{Q}|}
\sum_{t\in\mathcal{Q}}
\operatorname{Mean}
\left(
\left|
\mathbf{R}_{l,t+q}[:,S_{t+q-1}:]
-
\mathbf{R}_{l,t}[:,S_{t-1}:]
\right|
\right).
\label{eq:direction_vector_sample}
\end{equation}
Here, $\operatorname{Mean}(\cdot)$ denotes averaging over the column dimension corresponding to token positions, and $\mathbf{v}_l\in\mathbb{R}^{d}$ represents the contribution change at layer $l$ averaged across all comparison pairs.

We then aggregate the sample-level vectors across all attack calibration samples under the same request condition to obtain the attack TRC score vector:
\begin{equation}
\mathscr{T}^{\mathrm{A}}_l
=
\frac{1}{|\mathcal{D}_{\mathrm{A}}|}
\sum_{z\in\mathcal{D}_{\mathrm{A}}}
\mathbf{v}_l(z),
\label{eq:TRC_score}
\end{equation}
$\mathbf{v}_l(z)$ is the contribution-change vector computed for sample $z$.
The resulting $\mathscr{T}^{\mathrm{A}}_l\in\mathbb{R}^{d}$ represents the attack TRC scores at layer $l$, with each entry corresponding to one hidden coordinate.

We analogously aggregate the contribution-change vectors over the benign calibration set to obtain $\mathscr{T}^{\mathrm{N}}_l$.
For layer-wise analysis, we summarize the attack and benign TRC scores over hidden coordinates.
For a controllable window span $h\in\{1,\ldots,L-1\}$, we define the normal-relative magnitude and the cross-layer variation as:
\begin{equation}
a_l=\frac{\operatorname{Mean}(\mathscr{T}^{\mathrm{A}}_l)}{\operatorname{Mean}(\mathscr{T}^{\mathrm{N}}_l)},
\qquad
e_l^A=\left(\frac{\operatorname{Mean}(\mathscr{T}^{\mathrm{A}}_{l+h})-\operatorname{Mean}(\mathscr{T}^{\mathrm{A}}_l)}{h}\right)^2.
\label{eq:trc_layer_statistics}
\end{equation}
Where, $a_l$ measures the magnitude of attack-induced contribution changes relative to benign generation, while
$e_l^A$ measures the net variation of the attack TRC scores over a local layer interval of span $h$.
Attention and MLP modules are calibrated separately within each model.

To normalize the cross-layer variation using benign observations only, we construct a benign layer-wise reference curve from $\mathcal{D}_{\mathrm{N}}$ using the same aggregation procedure as for the attack samples.
We collect its positive cross-layer variation values and define the TRC-l score as
\begin{equation}
\mathscr{L}_l
=
a_l
\left(
1+
\frac{
e_l^{\mathrm{A}}
}{
\exp\left(
\operatorname{Mean}_{e\in\mathcal{E}_{\mathrm{N}}}
\log e
\right)
}
\right),
\qquad
\mathcal{E}_{\mathrm{N}}
=
\left\{
e_l^{\mathrm{N}}
\mid
0\leq l\leq L-h-1,
\ e_l^{\mathrm{N}}>0
\right\}.
\label{eq:trc_l_score}
\end{equation}
$e_l^{\mathrm{N}}$ denotes the cross-layer variation of the benign reference curve starting at layer $l$, computed using the same formulation as $e_l^{\mathrm{A}}$.
The geometric mean over $\mathcal{E}_{\mathrm{N}}$ provides a benign reference scale for the variation term, such that $\mathscr{L}_l$ jointly captures the normal-relative magnitude and the normalized cross-layer variation of the attack curve.

We use $\mathscr{L}_l$ as the criterion for localizing the intervention layer, favoring candidate starting layers with both a low normal-relative magnitude and limited variation over the following $h$ layers.
The starting layer is selected by minimizing $\mathscr{L}_l$, with ties broken in favor of the shallowest layer
$l^\star=\min\left(\operatorname*{argmin}_{0\leq l\leq L-h-1}\mathscr{L}_l\right)$.
The selected localization window is represented by $(l^\star,h)$, where $l^\star$ specifies the starting layer and $h$ is the controllable window span.

\subsection{Selective Residual-Stream Intervention}
\label{sec:method_suppression}

The localization stage returns $l^\star$ as the selected intervention layer.
At $l^\star$, we use the TRC scores computed from attack samples $\mathscr{L}^{\mathrm{A}}_{l^\star}$, to identify the residual-stream contributions most associated with repetition.
Given a direction-selection ratio $\rho\in(0,1]$, we select
$\Omega_{l^\star}=\operatorname{Top}_{\lfloor \rho d \rfloor}\left(\mathscr{T}^{\mathrm{A}}_{l^\star}\right)$,
where $\Omega_{l^\star}$ contains the $\lfloor \rho d \rfloor$ coordinates with the largest attack TRC scores at the selected layer.
We further determine the suppression strength directly from the layer-level TRC-l score:
\begin{equation}
\alpha_{l^\star}
=
\frac{
\max\left(
0,
\ln \mathscr{L}^{\mathrm{N}}_{l^\star}
-
\ln \mathscr{L}^{\mathrm{A}}_{l^\star}
\right)
}{
1+
\max\left(
0,
\ln \mathscr{L}^{\mathrm{N}}_{l^\star}
-
\ln \mathscr{L}^{\mathrm{A}}_{l^\star}
\right)
}.
\label{eq:adaptive_suppression_strength}
\end{equation}
The resulting $\alpha_{l^\star}\in[0,1)$ increases as the TRC-l score decreases, assigning stronger suppression to layers exhibiting a smaller normal-relative magnitude and more stable cross-layer variation.
The direction-selection ratio $\rho$ controls the intervention sparsity, while $\alpha_{l^\star}$ controls the suppression strength.
During inference, let $\mathbf{B}_{l^\star}\in\mathbb{R}^{d\times S}$ denote the current module contribution over $S$ sequence positions and $\mathbf{U}_{l^\star}\in\mathbb{R}^{d\times S}$ the residual stream entering the corresponding residual addition.
We construct a diagonal soft mask $\mathbf{M}_{l^\star}\in[0,1]^{d\times d}$ and apply it in residual addition:
\begin{equation}
[\mathbf{M}_{l^\star}]_{u,u}=
\begin{cases}
1-\alpha_{l^\star}, & u\in\Omega_{l^\star},\\
\end{cases}
\qquad
\widetilde{\mathbf{B}}_{l^\star}
=
\mathbf{B}_{l^\star}+\mathbf{M}_{l^\star}\mathbf{U}_{l^\star}.
\label{eq:masked_residual_addition}
\end{equation}
Left-multiplying $\mathbf{U}_{l^\star}$ by $\mathbf{M}_{l^\star}$ scales each selected row by $1-\alpha_{l^\star}$ in residual addition.
The mask is shared across sequence positions and is applied only to the selected target layer, leaving all other layers unchanged.
For every layer $l\neq l^\star$, the module contribution is added without modification.
The selected layer, coordinate set, and soft mask are fixed after calibration and directly applied to subsequent evaluation requests.

\section{Experiments}
\label{sec:experiments}

\subsection{Experimental Setup}
\label{sec:experiments_setup}

\paragraph{Models and Evaluation Scope}
\label{sec:experiments_setup_models}
Our evaluation suite comprises three groups. The LVLM group includes InstructBLIP-Vicuna-7B~\citep{instructblip}, Qwen2.5-VL-3B-Instruct~\citep{qwen2.5-VL}, and LLaVA-1.5-7B~\citep{Liu_2024_CVPR}. The text-only LLM group includes Llama-3.2-3B\footnote{Official Llama 3.2 model card: \url{https://github.com/meta-llama/llama-models/blob/main/models/llama3_2/MODEL_CARD.md}.} and Qwen2.5-3B~\citep{qwen2.5}. The reasoning-model (LRM) group includes DeepSeek-Llama-8B~\citep{deepseekai2025deepseekr1incentivizingreasoningcapability}, Qwen3.6-27B~\citep{qwen3.6-27b}, and GLM-4.7-Flash~\citep{5team2025glm45agenticreasoningcoding}. 


\paragraph{Attack and Benign Evaluation Data}
\label{sec:experiments_setup_data}
The attack suite separates visual perturbations from textual induction. RECITE optimizes image perturbations to elicit repeated output~\citep{gao2025resourceconsumptionredteaminglarge}. Textual conditions comprise GCG-based prompt optimization~\citep{zou2023universaltransferableadversarialattacks}, LoopLLM's repetition-inducing attacks~\citep{Li_2026}, and direct repetition instructions (Direct). 
Benign evaluation uses ScienceQA for multimodal science question answering~\citep{NEURIPS2022_11332b6b} and TextVQA for answering questions requiring scene-text understanding~\citep{Singh_2019_CVPR}. MMLU assesses knowledge and reasoning across text-based subjects for the LLMs and LRMs~\citep{hendryckstest2021}.
Table~\ref{tab:evaluation_scope} in the Appendix specifies the attack coverage and benign tasks for each model.

\paragraph{Baselines}
\label{sec:experiments_setup_baselines}
We compare TRC with three defense baselines, alongside the unmodified model.
\textit{Fixed-length truncation (Fixed-length)} stops decoding at a preset output-token limit~\citep{zhang-etal-2025-pd3f,gao2025resourceconsumptionredteaminglarge}.
\textit{No-repeat} uses no-repeat $n$-gram blocking to prevent any next token from reproducing a previously generated token $n$-gram~\citep{ICLR2026_8d321ebb, zhu-etal-2023-penalty}.
\textit{Interpretability-based intervention (AUSteer)} adapts AUSteer's selection and steering of individual activation dimensions to repetition suppression~\citep{ICLR2026_423d0909}.

\paragraph{Evaluation Metrics}
\label{sec:experiments_setup_metrics}
We evaluate generation length, task accuracy, and loop rate.
Following Hiraoka and Inui~\citep{hiraoka-inui-2025-repetition}, we identify repetitive outputs by detecting recurring token subsequences in the generated sequence. The loop rate is defined as $N_{\mathrm{rep}}/N$, where $N_{\mathrm{rep}}$ is the number of samples exhibiting repetitive loops and $N$ is the total number of evaluated samples.


\subsection{Mitigation Effectiveness and Generalization}
\label{sec:experiments_defense}
\paragraph{Defense Effectiveness.}
\begin{table}[t]
\centering
\vspace{-15pt}
\caption{Defense results on three LVLMs. Lower is better for both metrics.}
\label{tab:main_def_lvlm}
\resizebox{\textwidth}{!}{%
\begin{tabular}{@{}ll|cc|cc|cc@{}}
\toprule
 & & \multicolumn{2}{c|}{\textbf{RECITE (visual)}} & \multicolumn{2}{c|}{\textbf{GCG (textual)}} & \multicolumn{2}{c}{\textbf{LoopLLM (textual)}} \\ \cmidrule(l){3-8}
\textbf{Model} & \textbf{Defense} &\quad \textbf{Length} \quad\quad&\quad \textbf{Loop rate} \quad\quad&\quad \textbf{Length} \quad\quad&\quad \textbf{Loop rate} \quad\quad&\quad \textbf{Length} \quad\quad&\quad \textbf{Loop rate} \quad\quad\\ \midrule
\rowcolor[HTML]{EFEFEF}
\cellcolor[HTML]{FFFFFF}InstructBLIP-7B & No defense & 1,866.44 & 92.0\% & 1,531.56 & 96.0\% & 1,833.96 & 100.0\% \\
 & Fixed-length & 389.84 & 92.0\% & 1,017.76 & 96.0\% & 952.36 & 88.0\% \\
 & No-repeat & 18.36 & 4.0\% & 17.88 & 4.0\% & 22.04 & 0.0\% \\
 & AUSteer & 413.16 & 20.0\% & 85.40 & 4.0\% & 165.84 & 8.0\% \\
\rowcolor[HTML]{D0D0D0}
\cellcolor[HTML]{FFFFFF} & \textbf{TRC} & 422.04 & 16.0\% & 1,312.56 & 56.0\% & 1,229.72 & 60.0\% \\ \midrule
\rowcolor[HTML]{EFEFEF}
\cellcolor[HTML]{FFFFFF}Qwen2.5-VL-3B & No defense & 4,096.00 & 100.0\% & 4,096.00 & 100.0\% & 1,395.32 & 32.0\% \\
 & Fixed-length & 62.00 & 100.0\% & 1,241.00 & 100.0\% & 177.16 & 32.0\% \\
 & No-repeat & 40.08 & 0.0\% & 35.28 & 4.0\% & 29.80 & 0.0\% \\
 & AUSteer & 4,096.00 & 76.0\% & 4,096.00 & 100.0\% & 4,096.00 & 84.0\% \\
\rowcolor[HTML]{D0D0D0}
\cellcolor[HTML]{FFFFFF} & \textbf{TRC} & 62.44 & 0.0\% & 2,469.72 & 60.0\% & 1,044.28 & 28.0\% \\ \midrule
\rowcolor[HTML]{EFEFEF}
\cellcolor[HTML]{FFFFFF}LLaVA-7B & No defense & 3,611.08 & 88.0\% & 3,113.60 & 76.0\% & -- & -- \\
 & Fixed-length & 181.72 & 88.0\% & 924.12 & 76.0\% & -- & -- \\
 & No-repeat & 34.80 & 0.0\% & 7.04 & 0.0\% & -- & -- \\
 & AUSteer & 1,250.32 & 60.0\% & 715.32 & 32.0\% & -- & -- \\
\rowcolor[HTML]{D0D0D0}
\cellcolor[HTML]{FFFFFF} & \textbf{TRC} & 362.56 & 8.0\% & 4.84 & 0.0\% & -- & -- \\ \bottomrule
\end{tabular}%
}
\end{table}

\begin{wraptable}{r}{0.40\textwidth}
\vspace{-0.75\baselineskip}
\centering
\caption{Benign task accuracy (\%).}
\setlength{\tabcolsep}{3pt}
\resizebox{0.39\textwidth}{!}{%
\begin{tabular}{@{}l|cc|c@{}}
\toprule
Method & \multicolumn{1}{c}{ScienceQA} & \multicolumn{1}{c|}{TextVQA} & \multicolumn{1}{c}{Average} \\ \midrule
\rowcolor[HTML]{EFEFEF}
Benign & 82.00\% & 79.00\% & 80.50\% \\
Fixed-length & 82.00\% & 79.00\% & 80.50\% \\
No-repeat & 82.00\% & 78.50\% & 80.25\% \\
AUSteer & 76.00\% & 60.50\% & 68.25\% \\
\rowcolor[HTML]{C0C0C0}
TRC & 81.50\% & 78.50\% & 80.00\% \\ \bottomrule
\end{tabular}%
}
\label{tab:main_def_nor_lvlm}
\vspace{-0.5\baselineskip}
\end{wraptable}

Table~\ref{tab:main_def_lvlm} shows that TRC consistently suppresses uncontrolled repetition across visual and textual attacks, reducing both generation length and loop rate in most settings. 
Its numerical efficiency is weaker than direct decoding constraints such as no-repeat, since TRC intervenes through localized internal signals rather than explicitly blocking repeated output patterns. 
Importantly, Table~\ref{tab:main_def_nor_lvlm} shows that this intervention has little effect on benign task accuracy. These results indicate that TRC can mitigate repetition while preserving normal generation, supporting the identified residual changes as meaningful intervention targets.

\begin{figure}[t]
    \centering
    \includegraphics[width=\textwidth]{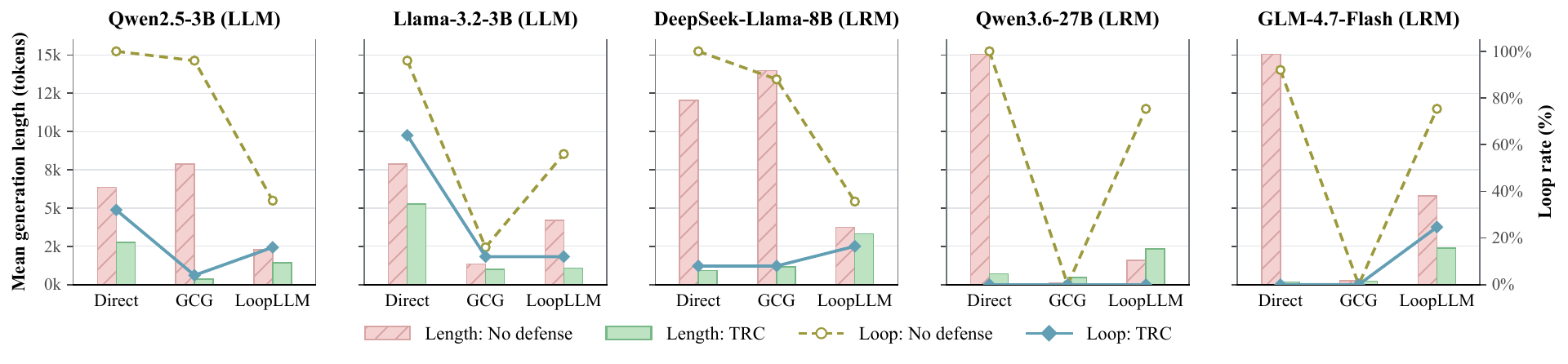}
    \vspace{-23pt}
    \caption{Generalization of TRC to LLMs and LRMs under Direct, GCG, and LoopLLM attacks. Bars show generation length and lines show loop rate; lower is better for both.}
    \label{fig:generalization_llm_lrm}
\end{figure}

\paragraph{Generalization.}
\label{sec:experiments_generalization}
We next examine whether TRC generalizes beyond LVLMs to text only LLMs and LRMs. 
As shown in Figure~\ref{fig:generalization_llm_lrm}, TRC consistently reduces generation length and loop rate across model families whenever the underlying attack succeeds, with especially strong effects on LRMs. 
Table~\ref{tab:main_def_nor_llm_lrm} shows that these gains come with little change in benign MMLU accuracy. 
We further observe more stable intervention effects on newer and larger models. A possible explanation is that advances in model scale and architecture lead to better separation between repetition related residual changes and normal semantic representations, allowing targeted suppression to preserve task relevant behavior more effectively.
The cross model results indicate that TRC captures internal repetition signals that transfer beyond the original LVLM setting.


\begin{table}[t]
\centering
\caption{Benign task accuracy on LLMs and LRMs. TRC causes  minor changes in performance.}
\label{tab:main_def_nor_llm_lrm}
\setlength{\tabcolsep}{4pt}
\resizebox{\textwidth}{!}{%
\begin{tabular}{@{}l|cc|ccc@{}}
\toprule
 & \multicolumn{2}{c|}{\textbf{LLM}} & \multicolumn{3}{c}{\textbf{LRM}} \\ \cmidrule(l){2-6}
\textbf{Setting} \quad\quad\quad\quad\quad\quad& \textbf{Qwen2.5-3B} & \textbf{Llama-3.2-3B} & \textbf{DeepSeek-Llama-8B} & \textbf{Qwen3.6-27B} & \textbf{GLM-4.7-Flash} \\ \midrule
\rowcolor[HTML]{EFEFEF}
Benign & 61.0\% & 55.0\% & 97.0\% & 99.5\% & 100.0\% \\
\rowcolor[HTML]{D0D0D0}
\textbf{TRC} & 61.0\% & 56.0\% & 95.0\% & 99.5\% & 100.0\% \\
Difference (pp) & 0.0\% & +1.0\% & -2.0\% & 0.0\% & 0.0\% \\ \bottomrule
\end{tabular}%
}
\end{table}


\subsection{Mechanistic Analysis}
\label{sec:experiments_mechanism}

We use Qwen2.5-VL-3B-Instruct as the primary model for the detailed mechanistic analyses in this section; experiments comparing model families include the other specified models.

\begin{figure*}[t]
    \centering
    \begin{minipage}[t]{0.50\textwidth}
        \centering
        \includegraphics[width=\linewidth]{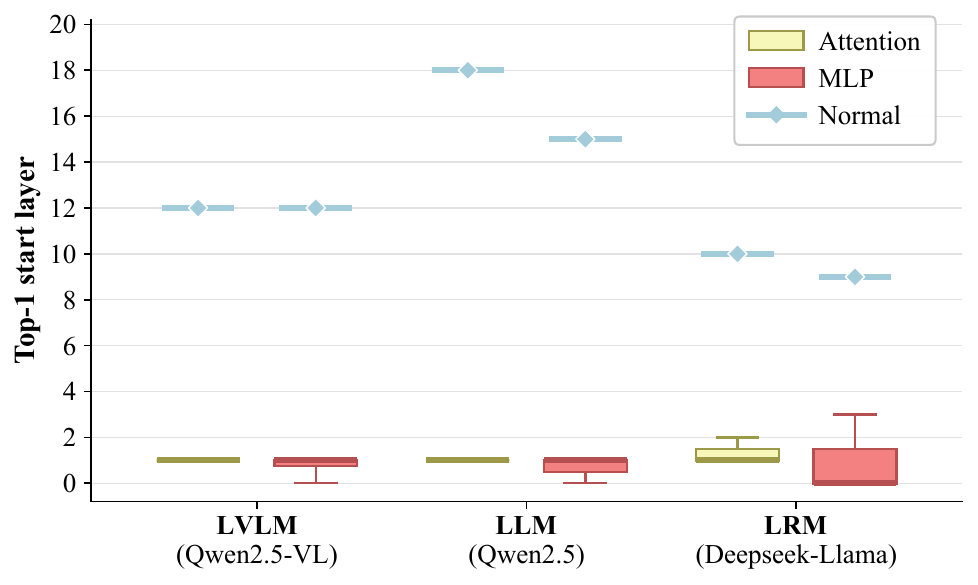}
        \\(a) Top 1 TRC localization across model families.
    \end{minipage}
    \hfill
    \begin{minipage}[t]{0.48\textwidth}
        \centering
        \includegraphics[width=\linewidth]{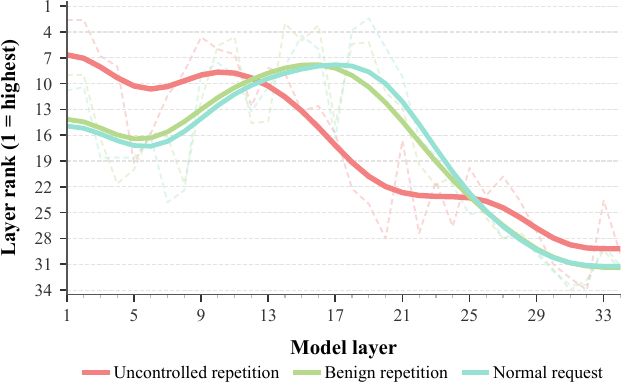}
        \\(b) Cross layer ranking of TRC
    \end{minipage}
    \caption{Layer wise localization of uncontrolled repetition. (a) Top 1 TRC localization across LVLM, LLM, and LRM families, showing consistently shallow attack locations. (b) Cross layer TRC rankings under uncontrolled repetition, benign repetition, and normal requests, revealing a distinct early layer profile for uncontrolled repetition.}
    
    \label{fig:mechanistic_localization}
\end{figure*}

\paragraph{Shallow Localization of Repetition Signals.}
\label{sec:experiments_mechanism_localization}
Figure~\ref{fig:mechanistic_localization}(a) compares the Top-1 layers identified by TRC for an LVLM, a text-only LLM, and an LRM. Across all evaluated attack conditions, both Attention and MLP branches consistently localize repetition signals to shallow layers within 0--3, whereas benign references are localized substantially later, between layers 9 and 18. 
This separation is preserved across three models, suggesting that uncontrolled repetition residual changes emerge early rather than at a model-specific depth. Attention provides the more stable localization signal, with nine of ten conditions concentrated at layer 1, while MLP locations vary across layers 0, 1, and 3. 
These results therefore identify the residual stream entering shallow Attention layers as a particularly consistent location of repetition related changes. Notably, its concentration at layer 1 corresponds to the residual representation immediately after the layer 0 MLP, indicating that repetition related changes are already prominent after the first MLP transformation.


\paragraph{Specificity to Uncontrolled Repetition.}
\label{sec:experiments_mechanism_specificity}
To test whether TRC merely responds to repeated tokens or repeated semantics, we compare layer-wise localization under three conditions: uncontrolled-repetition failures, benign requests containing legitimate repetition, and ordinary normal requests. 
Figure~\ref{fig:mechanistic_localization}(b) shows that benign repetition closely follows the localization profile of normal requests across layers. In both cases, the highest-ranked region shifts toward the middle layers before weakening in later layers. 
Uncontrolled repetition follows a different trajectory, with candidate locations ranked substantially higher in the early layers and progressively lower at greater depths.
This separation rules out the simple explanation that TRC detects repetition semantics or repeated token identity alone. Instead, the localized signal is associated with uncontrolled repetition collapse. 
The result therefore supports the specificity of TRC to failure-related repetition dynamics, while the causal role of the localized components is evaluated separately in subsequent chapters.
The absolute scores and their depth-dependent trends are examined in Appendix~\ref{app:layerwise_trc_profiles}.


\paragraph{Attention and MLP Branch Contributions.}
\label{sec:experiments_mechanism_branches}

\begin{figure*}[t]
    \centering
    \begin{minipage}[t]{0.50\textwidth}
        \centering
        \includegraphics[width=\linewidth]{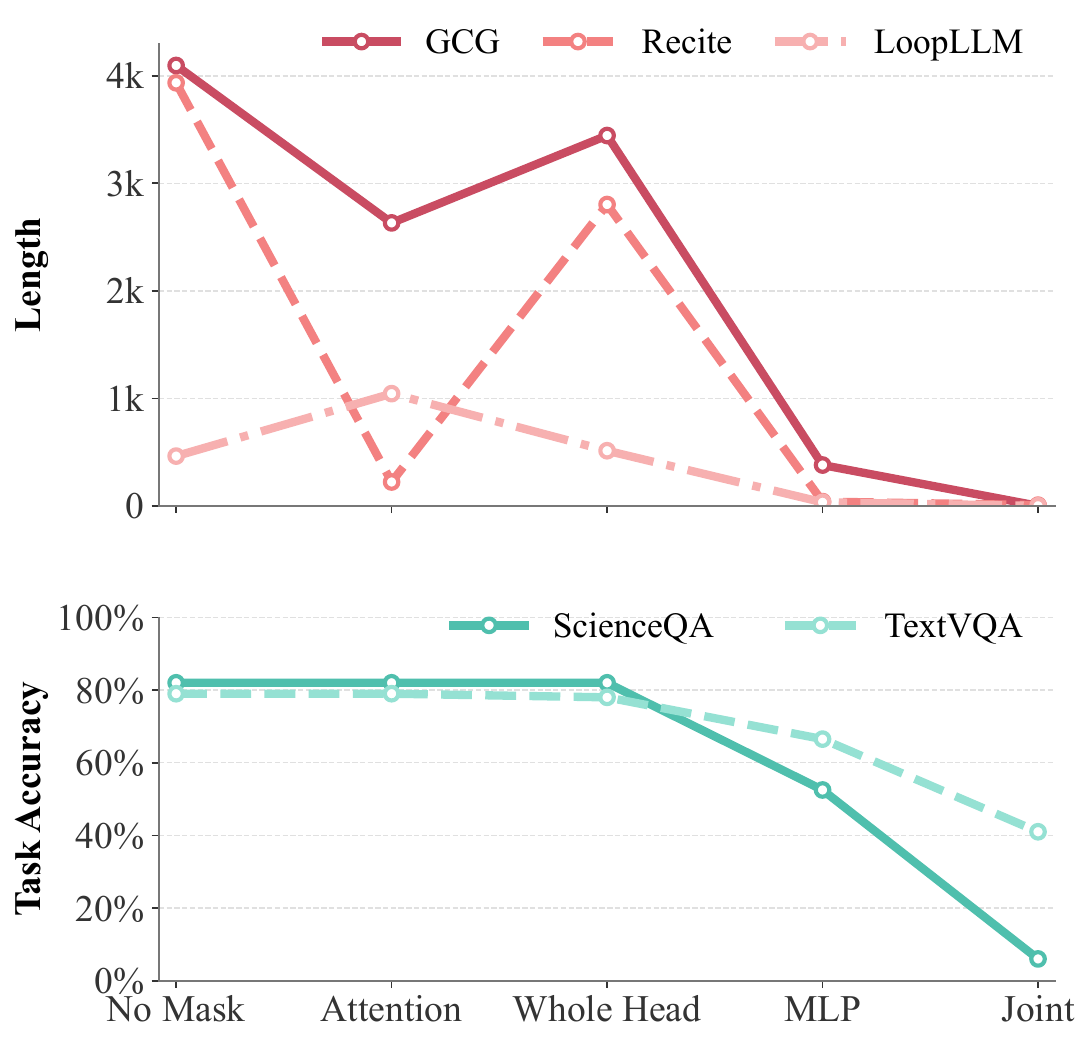}
        \\(a) Branch selection
    \end{minipage}
    \hfill
    \begin{minipage}[t]{0.48\textwidth}
        \centering
        \includegraphics[width=\linewidth]{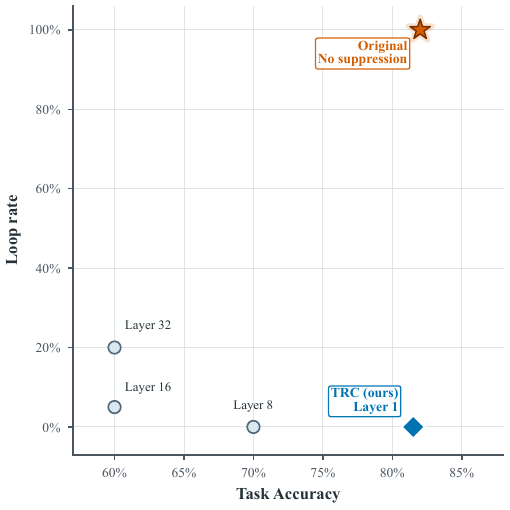}
        \\(b) Layer selection
    \end{minipage}
    \caption{Mechanistic evidence for branch and layer selection.
    (a) Branch-wise suppression on attack and normal examples. Attention-only suppression preserves normal accuracy, whereas MLP-only and joint suppression reduce attack length at substantially higher utility cost.
    (b) Effect of the suppression start layer on Recite repetition suppression and ScienceQA accuracy. The star denotes the original model without suppression.}
    \label{fig:branch_contributions}
    \label{fig:layer_suppression_tradeoff}
\end{figure*}

Figure~\ref{fig:branch_contributions}(a) shows that suppressing the full Attention write provides a better mitigation–utility tradeoff than suppressing individual heads, suggesting that repetition is not yet concentrated in a specific head at shallow layers. In contrast, suppressing the MLP write or both branches causes substantially larger degradation on benign tasks. Since the intervened residual stream already contains the output of the preceding block, these results suggest that repetition related features are formed early in MLP computations and then propagated through subsequent residual updates. This interpretation is consistent with prior studies characterizing MLPs as key value memories and linking them to concept and knowledge representations \citep{geva-etal-2021-transformer,geva-etal-2022-transformer,dai-etal-2022-knowledge,NEURIPS2022_6f1d43d5}.



\paragraph{Intervention Effects of Layer and Direction Selection.}
\label{sec:experiments_mechanism_causal}
To isolate the effect of layer choice, we keep the suppression rule fixed and vary only the intervention layer. As shown in Figure~\ref{fig:layer_suppression_tradeoff}(b), the shallow layer selected by TRC achieves the best mitigation--utility trade-off: layer 1 suppresses all of repetitive failures while retaining 81.5\% ScienceQA accuracy, close to the original model. Applying the same intervention at deeper layers can maintain high suppression but causes substantially larger accuracy degradation. These results show that effective repetition mitigation depends critically on where suppression is applied, supporting TRC's shallow-layer localization rather than depth-agnostic intervention.


\paragraph{Attention Update Magnitude and Residual Propagation.}
\label{sec:experiments_mechanism_residual_propagation}
\begin{wraptable}{r}{0.40\textwidth}
    \vspace{-20pt}
    \centering
    \caption{Attention-update magnitudes in the first two blocks. Gap/$\tau_l<1$ indicates that the condition difference remains within the normal token-level fluctuation bound.}
    \label{tab:attention_update_magnitude}
    \scriptsize
    \setlength{\tabcolsep}{2.5pt}
    \resizebox{\linewidth}{!}{%
        \begin{tabular}{@{}clrrr@{}}
            \toprule
            \textbf{Block} & \textbf{Metric} & \textbf{Normal} & \textbf{Repeat} & \textbf{Gap/$\tau_l$} \\
            \midrule
            0 & L2  & 22.2009 & 22.8838 & 0.234 \\
            0 & RMS &  0.4946 &  0.5057 & 0.172 \\
            \midrule
            1 & L2  & 12.9280 & 13.7456 & 0.730 \\
            1 & RMS &  0.2867 &  0.3038 & 0.692 \\
            \bottomrule
        \end{tabular}%
    }
    \vspace{-0.8em}
\end{wraptable}
We further examine whether shallow attention blocks repeatedly amplify the repetition signal, or whether the signal is mainly retained through residual propagation. For the first two attention blocks, we measure the update magnitude $\mathbf{D}_l=\mathbf{U}^{\mathrm{out}}_l-\mathbf{U}^{\mathrm{in}}_l$ using both L2 norm and RMS, and normalize the condition gap by the corresponding normal fluctuation threshold $\tau_l$. As shown in Table~\ref{tab:attention_update_magnitude}, the repetition-induced gaps remain within normal fluctuation ranges for both metrics, reaching only 23.4\%/17.2\% of the threshold in Block 0 and 73.0\%/69.2\% in Block 1 for L2/RMS, respectively. This indicates that repetitive generation is not accompanied by an abnormal increase in shallow attention-update magnitude. Combined with the earlier localization results, this is more consistent with repetition-related features emerging through shallow attention transformations and then being retained through the residual stream, rather than being repeatedly amplified by subsequent attention transformations. This early emergence precedes the progressive strengthening of repetition related semantics in intermediate layers, as observed in prior activation level studies \citep{hiraoka-inui-2025-repetition,yao-etal-2025-understanding}.

\paragraph{Propagation and Activation Restoration.}
\label{sec:experiments_mechanism_propagation}
We test whether the coordinates localized by TRC capture an early attack-associated feature using GCG examples that differ only by the optimized suffix. 
At the first layer, we perform unit-strength cross-interventions by either restoring localized attack coordinates with values from the paired normal trajectory or injecting the corresponding attack-minus-normal difference into the normal trajectory, with equal-size random coordinates as controls. 
As shown in Table~\ref{tab:cross_intervention}, restoring the TRC-localized coordinates converts 80\% repetitive attack trajectories to non-repetitive outputs and reduces average generation length from 4096 to 828 tokens, substantially outperforming random restoration. 
This indicates that the localized coordinates already encode behaviorally relevant attack-associated information at the first layer and that restoring these early activations can propagate to downstream recovery. 
Conversely, injecting the attack-associated difference into normal trajectories disrupts generation but does not reproduce repetition, suggesting that these coordinates are important for the failure state but are not independently sufficient to induce it. Together, the results support TRC as identifying early residual features that contribute to repetition and can be causally restored to mitigate the failure.
Appendix~\ref{app:cross_intervention_examples} presents representative outputs for both intervention directions.

\begin{table*}[t]
    \centering
    \vspace{-20pt}
    \caption{Cross intervention results with unit strength on GCG examples. \textit{TRC} denotes coordinates selected by TRC, while \textit{Random} uses an equal size coordinate set. Transition success measures repetition to nonrepetition for attack inputs and nonrepetition to repetition for normal inputs. $\Delta$Len is computed relative to the corresponding no intervention baseline.}
    \label{tab:cross_intervention}
    \footnotesize
    \setlength{\tabcolsep}{3pt}
    \begin{tabular}{@{}llp{3.0cm}rrrrp{2.4cm}@{}}
        \toprule
        \textbf{Input}
        & \textbf{Coordinates}
        & \textbf{Residual intervention}
        & \shortstack{\textbf{Repeat}\\\textbf{rate (\%)}}
        & \shortstack{\textbf{Transition}\\\textbf{success (\%)}}
        & \shortstack{\textbf{Avg.}\\\textbf{length}}
        & \shortstack{\textbf{$\Delta$Len}\\\textbf{(\%)}}
        & \textbf{Observed change} \\
        \midrule
        Attack & -- & None
        & 100.0 & -- & 4096 & -- & Repetition baseline \\
        \rowcolor[HTML]{D0D0D0}
        Attack & \textbf{TRC (ours)} & Normal residual
        & \textbf{20.0} & \textbf{80.0} & \textbf{828} & \textbf{$-79.8$}
        & Strong restoration \\
        Attack & Random & Normal residual
        & 60.0 & 40.0 & 2462 & $-39.9$
        & Partial restoration \\
        \midrule
        Normal & -- & None
        & 0.0 & -- & 11 & -- & Normal baseline \\
        \rowcolor[HTML]{D0D0D0}
        Normal & \textbf{TRC (ours)} & Attack addition
        & 0.0 & 0.0 & 2 & --
        & Output failure without repetition \\
        Normal & Random & Attack addition
        & 0.0 & 0.0 & 11 & --
        & Nearly unchanged \\
        \bottomrule
    \end{tabular}
\end{table*}



Appendix~\ref{app:ablation_studies} further reports sensitivity analyses on the layer window span, the number of attack training examples, the suppression ratio, and the maximum repetition distance used by the repeated $n$ gram detector. These results examine the robustness of TRC to its main design choices.
\section{Conclusion}
\label{sec:conclusion}

We presented \textit{Tokenwise Residual Comparison} (TRC), a framework for identifying and mitigating uncontrolled repetition by comparing residual stream contributions across generation positions.
Rather than focusing only on prominent repetition representations in intermediate or later layers, TRC tracks how residual contributions evolve along the generated sequence and localizes fine grained repetition signals to specific layers and coordinates. 
Across LVLMs, text only LLMs, and LRMs, TRC consistently identifies shallow repetition signals and enables targeted suppression that reduces repetitive generation while largely preserving benign task performance. 
Mechanistic analyses further show that these signals emerge early, remain distinguishable from benign and legitimate repetition, and can be partially restored through localized activation replacement, supporting residual propagation as a plausible mechanism by which repetition related features persist through the network. 
Together, these findings shift the analysis of uncontrolled repetition from where strong repetition representations are observed to how they emerge early and become actionable, providing a finer grained perspective for understanding and mitigating resource consumption failures in autoregressive models.

\section*{AI use statement}
Generative AI tools were used to assist with literature retrieval, review manuscript formatting, and suggest caption and editorial revisions. The authors reviewed all AI-assisted outputs and suggestions and take full responsibility for the final text, data, results, and claims.

\section*{Ethics statement}

This work studies repetition-inducing attacks and a defense against them using
existing model and benchmark data. Attack procedures are reported to support
evaluation and defense research; they may also be misused to increase inference
costs or disrupt model services. We report defensive results alongside benign-task
performance to make this trade-off visible.

\subsection*{Reproducibility statement}

The method and evaluation metrics are described in Sections~\ref{sec:method}
and~\ref{sec:experiments_setup}; the models, attack conditions, and benign tasks
are listed in Appendix~\ref{app:evaluation_scope}. The appendix also documents
the construction and validation of the legitimate-repetition controls.

\bibliography{iclr2027_conference}
\bibliographystyle{iclr2027_conference}

\clearpage
\appendix

\section{Model and Attack Coverage}
\label{app:evaluation_scope}
Table~\ref{tab:evaluation_scope} maps the attack pools and benign evaluation
tasks to the models used in our experiments. We construct attack pools
separately for each checkpoint so that model-specific tokenization, chat
templates, and multimodal preprocessing are preserved. A candidate is retained
only when greedy decoding produces an uncontrolled one- or two-token cycle for
at least ten consecutive repetitions. A dash in the table therefore denotes a
condition outside the evaluated scope, rather than a failed attack.

\paragraph{Training--test separation and default configuration.}
The attack examples used to localize TRC are disjoint from all examples used
for evaluation. For each target model, we perform localization once using a
designated training attack: RECITE for LVLMs and GCG for LLMs and LRMs. We do
not retrain or retune TRC separately for every evaluation attack or dataset;
instead, the resulting model-specific configuration is applied directly to the
held-out attack pools summarized in Table~\ref{tab:evaluation_scope}. In the
main experiments, the adaptive rule yields a suppression strength of
approximately $\alpha_{l^\star}=0.73$ (73\%). We use a direction-selection
ratio of $\rho=5\%$ and a layer-window span of $h=3$.

\begin{table*}[htbp]
  \centering
  \caption{Model-specific attack coverage and benign evaluation tasks. Y marks
  an included attack condition; -- denotes a condition outside the evaluated
  scope.}
  \label{tab:evaluation_scope}
  \scriptsize
  \setlength{\tabcolsep}{6pt}
  \renewcommand{\arraystretch}{1.08}
  \resizebox{\linewidth}{!}{%
  \begin{tabular}{@{}lcccc l@{}}
    \toprule
    Model & RECITE & GCG & LoopLLM & Direct & Benign tasks \\
    \midrule
    \rowcolor{gray!15}\multicolumn{6}{@{}l}{\textit{LVLMs}} \\
    InstructBLIP-Vicuna-7B & Y & Y & Y & -- & ScienceQA, TextVQA \\
    Qwen2.5-VL-3B-Instruct & Y & Y & Y & -- & ScienceQA, TextVQA \\
    LLaVA-1.5-7B & Y & Y & -- & -- & ScienceQA, TextVQA \\
    \rowcolor{gray!15}\multicolumn{6}{@{}l}{\textit{LLMs}} \\
    Llama-3.2-3B & -- & Y & Y & Y & MMLU \\
    Qwen2.5-3B & -- & Y & Y & Y & MMLU \\
    \rowcolor{gray!15}\multicolumn{6}{@{}l}{\textit{LRMs}} \\
    DeepSeek-Llama-8B & -- & Y & Y & Y & MMLU \\
    Qwen3.6-27B & -- & Y & Y & Y & MMLU \\
    GLM-4.7-Flash & -- & Y & Y & Y & MMLU \\
    \bottomrule
  \end{tabular}%
  }
\end{table*}

\paragraph{RECITE.}
For the multimodal models, we follow the original RECITE construction
procedure~\citep{gao2025resourceconsumptionredteaminglarge}. Each source item
contains an image, its associated request, and a repeated-token target. A
targeted projected-gradient attack modifies the image while leaving the text
request fixed. We then run the corresponding LVLM on the adversarial image and
retain only candidates whose generated continuation satisfies the repetition
criterion above. This produces paired clean and adversarial images for the same
request and avoids changing the linguistic content of the input.

\paragraph{GCG.}
We implement GCG~\citep{zou2023universaltransferableadversarialattacks} with
\href{https://github.com/GraySwanAI/nanoGCG}{NanoGCG} and adapt its optimization
to the tokenizer, chat template, and cache representation of each model family.
Starting from the clean request pool associated with a target model, NanoGCG
optimizes a discrete suffix toward a repeated continuation. The resulting
suffix is transferred into the corresponding target input format and screened
again on the actual target checkpoint. Thus, the GCG datasets contain
model-specific request--suffix pairs rather than one universal suffix reused
across all models.

\paragraph{LoopLLM.}
We construct the LoopLLM pools with the same model-wise adaptation and transfer
protocol used for GCG, but replace the GCG target loss with LoopLLM's
repetition-inducing objective~\citep{Li_2026}. For each clean request, coordinate
optimization searches for a suffix that concentrates probability mass on a
short cyclic token pattern. The suffix is appended to the request under the
target model's own chat template, and the generated continuation is retained
only after target-model screening. This keeps the base request intact while
making the optimized suffix specific to the model and source pool.

\paragraph{Direct.}
Direct follows the repeated-token attack setting studied by
\citet{pmlr-v267-yona25a}. We first serialize the original request with the
model's chat template, then concatenate a short repeated token sequence to the
assistant-side generation prefix. Autoregressive decoding resumes from this
prefilled partial output, so the model receives the repetition as its own
unfinished continuation rather than as a new user instruction. We retain
examples only when the newly generated tokens continue into an uncontrolled
repetition cycle.

\paragraph{Benign task sets.}
For utility evaluation, LVLMs use ScienceQA and TextVQA, which test multimodal
science reasoning and question answering over scene text,
respectively~\citep{NEURIPS2022_11332b6b,Singh_2019_CVPR}. Text-only LLMs and
LRMs use MMLU across its subject domains~\citep{hendryckstest2021}. These benign
inputs preserve the original images, questions, and answer choices and receive
no attack suffix or assistant-side repetition prefix.
For each dataset, we use a subset of 200 samples for evaluation.

\section{Distinction from Existing Interpretability Methods}
\label{app:interpretability_comparison}

Table~\ref{tab:interpretability_comparison} compares TRC with representative
interpretability approaches along the three axes most relevant to our design:
the quantity being observed, the internal site being localized, and the model
modalities on which the method has been demonstrated. The comparison concerns
methodological scope rather than a shared numerical benchmark; the listed
approaches answer complementary questions and were not all designed to mitigate
uncontrolled repetition.

\begin{table*}[htbp]
  \centering
  \caption{Comparison with representative interpretability approaches. ``Scope''
  records the model families demonstrated in the cited work, rather than a
  theoretical restriction. TRC differs primarily in observing changes between
  cycle-aligned generation positions, localizing the corresponding pre-addition
  residual writes, and applying the same formulation across LVLMs, LLMs, and
  LRMs.}
  \label{tab:interpretability_comparison}
  \scriptsize
  \setlength{\tabcolsep}{3pt}
  \renewcommand{\arraystretch}{1.12}
  \begin{tabular}{@{}>{\raggedright\arraybackslash}p{0.14\textwidth}>{\raggedright\arraybackslash}p{0.27\textwidth}>{\raggedright\arraybackslash}p{0.22\textwidth}>{\raggedright\arraybackslash}p{0.12\textwidth}>{\raggedright\arraybackslash}p{0.19\textwidth}@{}}
    \toprule
    Approach & Primary observation & Localization target & Demonstrated scope & Principal distinction from TRC \\
    \midrule
    FFN vocabulary analysis~\citep{geva-etal-2021-transformer,geva-etal-2022-transformer}
    & Activations and vocabulary-space promotion produced by feed-forward updates
    & FFN memories, values, and layer-wise concept promotion
    & Text LMs
    & Explains stored or promoted concepts, but does not compare residual changes along a repetitive generation trajectory. \\

    Layer-wise lexical probing~\citep{liu-etal-2024-fantastic}
    & Probe recoverability of lexical-semantic information from hidden representations
    & Layers at which lexical semantics are most accessible
    & Generative text LMs
    & Measures information accessibility through an external readout rather than locating native residual coordinates for intervention. \\

    Repetition neurons~\citep{hiraoka-inui-2025-repetition}
    & Neuron-activation changes before and after the onset of repetition
    & Repetition-associated neurons in intermediate and final layers
    & Text LMs
    & Directly studies repetition, but uses activation amplitude around onset rather than cycle-aligned changes before residual addition. \\

    SAE repetition features~\citep{yao-etal-2025-understanding}
    & Logit-based layer screening followed by learned sparse-feature activations
    & SAE features within selected layers
    & Text LLMs
    & Resolves features with an auxiliary learned dictionary; TRC operates on native attention and MLP contribution coordinates. \\

    Cross-modal information flow~\citep{Zhang_2025_CVPR}
    & Answer-performance changes after blocking attention between image and question positions
    & Layered cross-token attention pathways for visual--linguistic integration
    & LLaVA-style MLLMs
    & Explains modality fusion within MLLMs, but is not a repetition-localization method and is not evaluated on text-only LLMs or LRMs. \\

    \rowcolor{gray!15}\textbf{TRC (ours)}
    & \textbf{Cycle-aligned changes in native attention and MLP contributions, normalized by benign magnitude and local cross-layer variation}
    & \textbf{A shallow pre-addition residual-write layer and its highest-scoring hidden coordinates}
    & \textbf{LVLMs, LLMs, and LRMs}
    & \textbf{Uses one modality-agnostic scoring and masking formulation without adding an auxiliary interpreter or network module.} \\
    \bottomrule
  \end{tabular}
\end{table*}

\paragraph{Observation dimension.}
Most neuron, probe, and feature based approaches inspect activation magnitude,
prediction attribution, recoverable content, or a learned feature basis at a
position or stage. TRC instead treats repetition as a temporal change pattern:
it aligns positions separated by the detected repetition cycle and measures how
the native attention and MLP contributions change across those positions. The
attack curve is then interpreted relative to benign generation and its local
cross-layer variation. This observation axis is suited to distinguishing a
continuation that becomes dynamically stationary from one that continues to
advance semantically.

\paragraph{Localization position.}
TRC localizes both a layer and residual coordinates at the module output before
residual addition, with attention and MLP analyzed separately. Our experiments
consistently place the repetition-associated low-change regime in shallow
layers, whereas normal generation reaches its minimum later. This gives TRC an
early and directly actionable intervention site, rather than requiring a probe,
an SAE dictionary, or a search over late output activations. The comparison
does not imply that shallow layers are universally optimal for every behavior;
it identifies the location supported for uncontrolled repetition under our
evaluated settings.

\paragraph{Cross-modal applicability.}
TRC observes the Transformer backbone's residual pathway after modality-specific
inputs have entered the autoregressive decoder. Consequently, the same score,
layer-selection rule, and soft-mask construction can be used for visually
induced repetition in LVLMs and textually induced repetition in LLMs and LRMs.
The resulting advantage is not merely support for multimodal inputs: it is a
shared internal analysis and intervention interface across visual and textual
attack constructions, as demonstrated by the model coverage and held-out
evaluations in this paper.

\section{Construction of Legitimate-Repetition Controls}
\label{app:legitimate_repetition_controls}

\paragraph{Models.}
For generation, we cap the number of newly generated tokens at 2,048 for InstructBLIP-Vicuna-7B (its default), 4,096 for the other LVLMs, 8,192 for all text-only LLMs, and 16,384 for all LRMs.

The specificity analysis in Section~\ref{sec:experiments_mechanism_specificity}
requires a control condition that contains intentional repetition without an
uncontrolled generation failure. We construct this condition using two design
principles rather than copying benchmark instances. MLissard controls
sequential complexity through repeated applications of simple rules, while
BIG-bench emphasizes explicitly specified, auditable tasks that probe distinct
model capabilities~\citep{bueno-etal-2024-mlissard,srivastava2023beyond}.
Following these principles, each of our image-grounded items requires the model
to first produce a normal semantic description and then execute a finite
repetition instruction.

Each prompt contains two ordered requirements for Qwen2.5-VL-3B-Instruct. The
model must first write one concise English sentence grounded in the input image
and must then repeat a designated common English token exactly ten times. We
select repeat units that map to one token both in isolation and with a leading
space under the model tokenizer. The repeated token is excluded from the
semantic description so that the descriptive and repetitive portions can be
validated separately. 

An item is accepted only when the output begins with an image-grounded
description, contains an expected visual keyword before the repeated segment,
omits the repeat unit from that semantic prefix, and ends with ten contiguous
copies of the same token ID. All five constructed items pass these criteria.
Table~\ref{tab:legitimate_repetition_examples} shows two representative
examples. They preserve a bounded, task-compliant repetition after a meaningful
visual response, in contrast to an uncontrolled loop that displaces the task
answer or fails to terminate.

\begin{table*}[ht]
  \centering
  \caption{Two representative legitimate-repetition controls for
  Qwen2.5-VL-3B-Instruct. Each prompt requires an image-grounded description
  followed by exactly ten copies of a designated token. Output text is
  reproduced verbatim; line breaks are normalized for table layout.}
  \label{tab:legitimate_repetition_examples}
  \footnotesize
  \setlength{\tabcolsep}{5pt}
  \renewcommand{\arraystretch}{1.15}
  \begin{tabular}{p{0.10\textwidth}p{0.43\textwidth}p{0.39\textwidth}}
    \toprule
    Image & Prompt & Generated output \\
    \midrule
    \includegraphics[width=\linewidth]{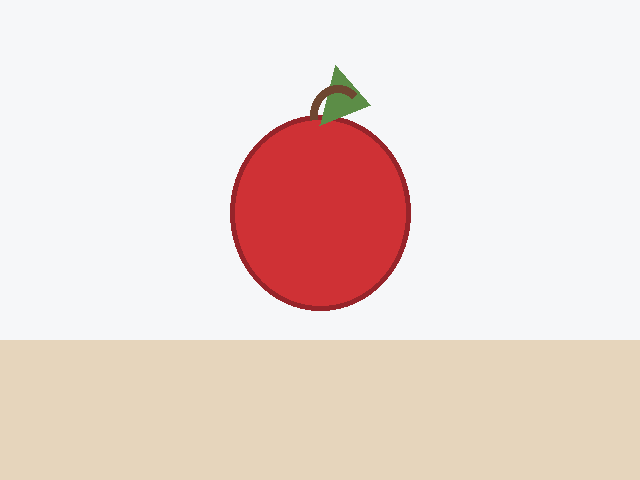}
    & Examine the image and return exactly two lines. Line 1 must begin with
      'Description:' and contain one concise factual English sentence describing
      the main visible object. Do not use the word 'echo' in line 1. Line 2 must
      be exactly: 'Repeat: echo echo echo echo echo echo echo echo echo echo.'
      Do not add any other text before, between, or after these two lines.
    & \begin{minipage}[t]{\linewidth}\raggedright
      \texttt{Description: The image depicts a simple, stylized red apple with
      a green leafy top.}\par
      \texttt{Repeat: echo echo echo echo echo echo echo echo echo echo}
      \end{minipage} \\
    \addlinespace
    \includegraphics[width=\linewidth]{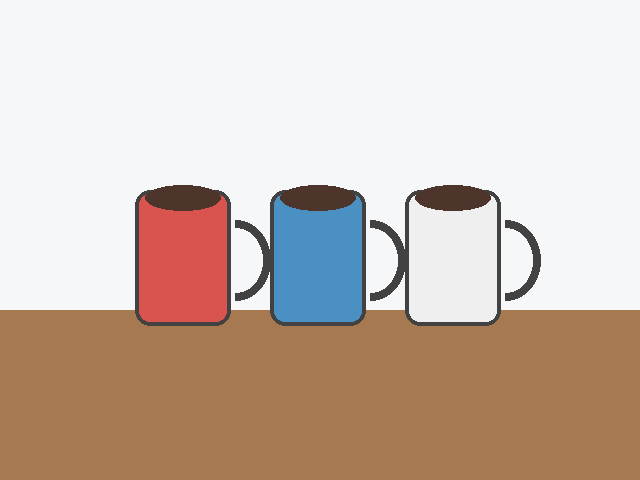}
    & Examine the image and return exactly two lines. Line 1 must begin with
      'Description:' and contain one concise factual English sentence describing
      the visible objects. Do not use the word 'signal' in line 1. Line 2 must be
      exactly: 'Repeat: signal signal signal signal signal signal signal signal
      signal signal.' Do not add any other text before, between, or after these
      two lines.
    & \begin{minipage}[t]{\linewidth}\raggedright
      \texttt{Description: Three mugs of different colors are lined up on a
      table.}\par
      \texttt{Repeat: signal signal signal signal signal signal signal signal
      signal signal.}
      \end{minipage} \\
    \bottomrule
  \end{tabular}
\end{table*}

\section{Layer-wise TRC-l Profiles under Repetition}
\label{app:layerwise_trc_profiles}

\begin{figure*}[t]
  \centering
  \includegraphics[width=0.98\textwidth]{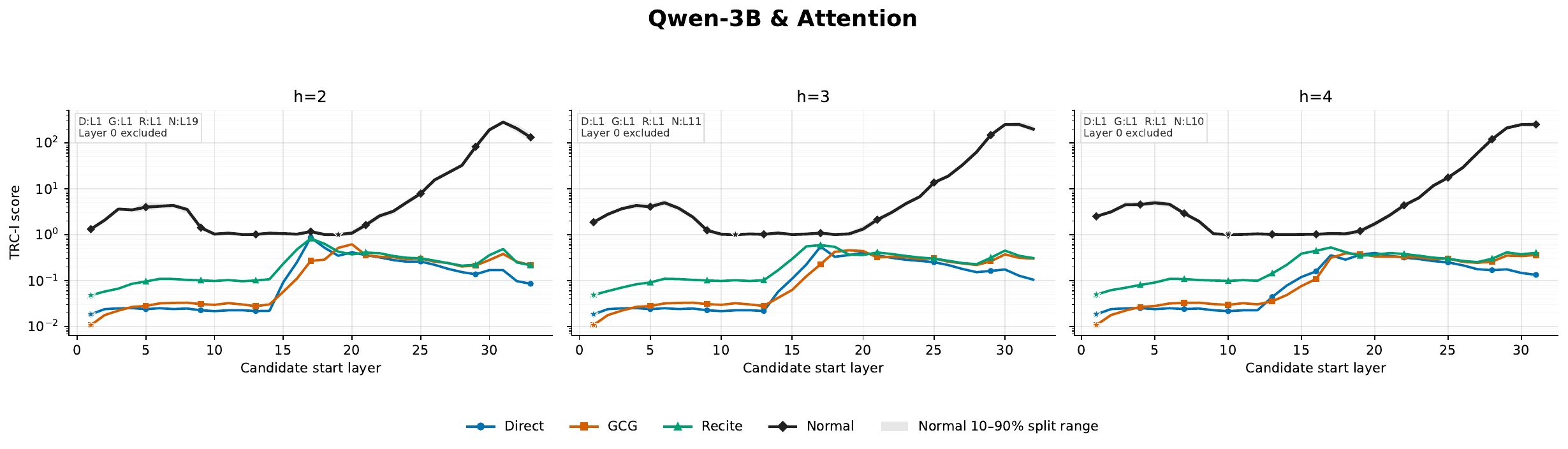}
  \par\smallskip
  \includegraphics[width=0.98\textwidth]{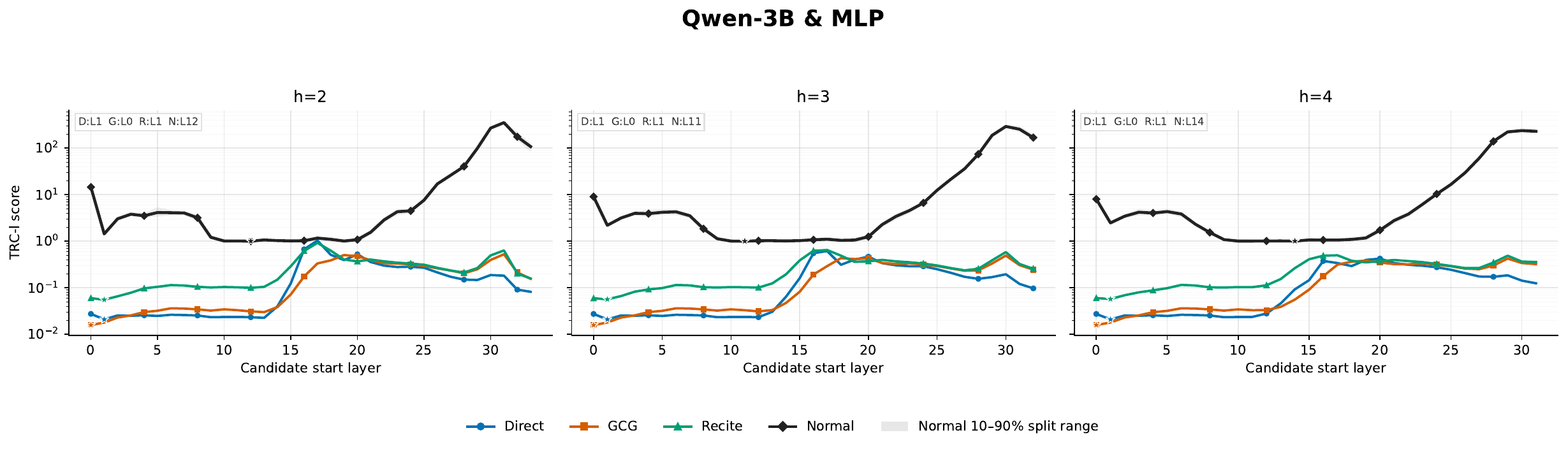}
  \caption{Layer-wise TRC-l profiles for Qwen2.5-VL-3B Attention (top) and MLP
  (bottom), with window spans $h=2,3,4$. Curves show Direct, GCG, and RECITE
  failures and the mean of normal-reference splits; gray shading denotes
  the normal range.}
  \label{fig:appendix_trc_l_profiles}
\end{figure*}

This analysis expands the shallow-localization and specificity results in
Sections~\ref{sec:experiments_mechanism_localization} and
\ref{sec:experiments_mechanism_specificity}. Figure~\ref{fig:appendix_trc_l_profiles}
plots the TRC-l score at every eligible candidate start layer for Direct, GCG,
RECITE, and normal requests on Qwen2.5-VL-3B. Attention and MLP are evaluated
separately at window spans $h\in\{2,3,4\}$. 
Lower scores indicate a smaller attack-to-normal contribution change ratio and/or less net change across the layer window, as defined in Section~\ref{sec:method_metric}. This does not affect layer ranking, and the score should not be interpreted as a direct measure of semantic similarity.

Across the plotted candidate layers, the failure curves generally lie below
the normal curve. This scale difference is compatible with the construction of
TRC: when successive comparison positions revisit a similar repetitive
content state, their aligned residual contributions can differ less than those
in a normal continuation that advances the answer. The smaller tokenwise
difference lowers the magnitude term of TRC-l. Because TRC-l also contains a
cross-layer variation term, and because comparison positions depend on the
detected repetition unit, the plot alone does not establish semantic similarity
as the unique cause of the score gap. The more diagnostic observation is where
each curve reaches its minimum within its own condition.

For normal requests, the minimum remains in the middle portion of the network:
Attention selects layers 19, 11, and 10 for $h=2,3,4$, respectively, while
MLP selects layers 12, 11, and 14. Ordinary generation can therefore exhibit
a relatively stable local contribution profile even while the model continues
to process a coherent task. This pattern is compatible with intermediate
layers integrating task-relevant information before later output decisions;
it does not establish a universal reasoning layer or imply that normal
representations are semantically unchanged between tokens.

The repetitive failures show a different location for their most stable,
low-change state. All three attacks select Attention layer 1 for every plotted
span. In MLP, Direct and RECITE select layer 1, whereas GCG selects layer 0.
The separation from normal requests persists for each tested $h$. Prior
layer-wise studies offer a more specific basis for interpreting this shallow
location: lower feed-forward layers tend to recognize relatively simple
textual patterns, whereas later layers capture more semantic patterns
\citep{geva-etal-2021-transformer}; probing a generative LLM likewise finds
lexical semantics in lower-layer representations
\citep{liu-etal-2024-fantastic}. Feed-forward updates can also promote
human-interpretable concepts in the model's vocabulary space
\citep{geva-etal-2022-transformer}. Together, these findings make it
plausible that basic lexical and local-pattern features in shallow layers
provide a substrate from which a repetition-associated representation can
emerge. Our TRC-l curves locate a low-change regime consistent with this
interpretation;
they do not identify a dedicated repetition neuron or prove that any cited
model implements the same failure mechanism.

Once established, such a feature can remain available to later blocks through
the residual stream, consistent with the account of feed-forward outputs
being refined across layers via residual connections
\citep{geva-etal-2021-transformer}. This interpretation also aligns with
Section~\ref{sec:experiments_mechanism_residual_propagation}: shallow
attention-update magnitudes are nonzero, yet their attack--normal gaps remain
within normal token-level fluctuation bounds. 

\section{Qualitative Cross-Intervention Examples}
\label{app:cross_intervention_examples}

We provide one representative paired example for each cross-intervention
direction discussed in Section~\ref{sec:experiments_mechanism_propagation}.
These examples illustrate the aggregate patterns in
Table~\ref{tab:cross_intervention}. In each pair, the normal and attack inputs share the same
base request, and the attack input differs only by the optimized GCG suffix.
The ellipsis below marks an output truncated for presentation.

\begin{center}
\fbox{%
\begin{minipage}{0.95\linewidth}
\small
\textbf{Example A: Restoring a normal activation on an attack trajectory.}

\smallskip
\textbf{Attack baseline}\quad
\texttt{Hello, hello, hello, hello, hello, ...}

\emph{Outcome:} 2048 newly generated tokens, repetitive, stopped at the
generation cap.

\smallskip
\textbf{Localized coordinates, normal residual replacement}\quad
\texttt{Hello, hello, hello, hello, hello.}

\emph{Outcome:} 11 newly generated tokens and non-repetitive generation.

\smallskip
\textbf{Matched random coordinates, normal residual replacement}
\quad \texttt{Hello, hello, hello, hello, hello, ...}

\emph{Outcome:} 2048 newly generated tokens, repetitive, stopped at the
generation cap.

\smallskip
\textbf{Observed contrast.} Restoring the localized coordinates recovers the
short normal response, whereas the equal-size random replacement leaves this
attack trajectory in the repetitive state.
\end{minipage}%
}
\end{center}

\begin{center}
\fbox{%
\begin{minipage}{0.95\linewidth}
\small
\textbf{Example B: Injecting the attack difference into a normal trajectory.}

\smallskip
\textbf{Normal baseline}\quad
\texttt{Hello, hello, hello, hello, hello!}

\emph{Outcome:} 11 newly generated tokens and non-repetitive generation.

\smallskip
\textbf{Localized coordinates, attack-minus-normal addition}
\quad \texttt{..}

\emph{Outcome:} 2 newly generated tokens and non-repetitive but erroneous
termination.

\smallskip
\textbf{Matched random coordinates, attack-minus-normal addition}
\quad \texttt{Hello, hello, hello, hello, hello!}

\emph{Outcome:} 11 newly generated tokens and non-repetitive generation,
matching the normal baseline.

\smallskip
\textbf{Observed contrast.} The localized perturbation severely disrupts the
normal response but does not recreate repetition; the matched random
perturbation leaves the response unchanged.
\end{minipage}%
}
\end{center}

\section{Ablation Studies}
\label{app:ablation_studies}

\subsection{Sensitivity to the Layer-Window Span}
\label{app:ablation_window_span}

The TRC-l score measures cross-layer variation over a local span $h$
(Section~\ref{sec:method_metric}). We vary $h\in\{2,3,4\}$ while keeping the
remaining localization procedure fixed. Table~\ref{tab:ablation_window_span}
reports the resulting Top-1 start layer for the attention and MLP branches.

\begin{table}[htbp]
  \centering
  \caption{Sensitivity of the localized Top-1 start layer to the layer-window
  span $h$. Each entry is Attention / MLP, and layers are zero-indexed.}
  \label{tab:ablation_window_span}
  \small
  \setlength{\tabcolsep}{10pt}
  \begin{tabular}{lccc}
    \toprule
    Attack & $h=2$ & $h=3$ & $h=4$ \\
    \midrule
    GCG     & $1/0$ & $1/0$ & $1/0$ \\
    RECITE  & $1/1$ & $1/1$ & $1/1$ \\
    LoopLLM & $1/1$ & $1/1$ & $1/1$ \\
    \bottomrule
  \end{tabular}
\end{table}

The selected locations are invariant over the three evaluated spans. In
particular, the branch-specific difference for GCG (attention layer 1 versus
MLP layer 0) is preserved, whereas both branches select layer 1 for RECITE and
LoopLLM.

\subsection{Sensitivity to the Suppression Ratio on Attacks}
\label{app:ablation_training_size}

This ablation varies the suppression ratio
$\rho\in\{1,5,10,15,20\}\%$ while keeping the attack training data and other
intervention settings fixed. Table~\ref{tab:ablation_training_size} reports
attack behavior; each attack cell gives mean newly generated tokens followed by
repetition rate. The final column reports the overlap between the coordinates
selected at each ratio and the coordinates selected at $\rho=20\%$.

\begin{table*}[t]
  \centering
  \caption{Sensitivity to the suppression ratio on attacks. Cells
  under each attack and the macro average report mean newly generated tokens /
  repetition rate (\%). Coordinate overlap is measured against the mask
  selected at $\rho=5\%$. Lower is better for both reported attack
  metrics.}
  \label{tab:ablation_training_size}
  \footnotesize
  \setlength{\tabcolsep}{5.5pt}
  \begin{tabular}{lccccc}
    \toprule
    $\rho$ (\%) & GCG & RECITE & LoopLLM & Macro avg. & Coordinate overlap \\
    \midrule
    No defense & $4096.00/100.0$ & $4096.00/100.0$ & $1395.32/32.0$ & $3195.77/77.3$ & -- \\
    1  & $1809.56/44.0$ & $54.80/0.0$  & $444.64/20.0$  & $769.67/21.3$  & $8/103$ \\
    5  & $2469.72/60.0$ & $62.44/0.0$ & $1044.28/28.0$ & $1192.15/29.3$ & $67/103$ \\
    10 & $2631.92/64.0$ & $222.60/4.0$ & $1044.24/28.0$ & $1299.59/32.0$ & $70/103$ \\
    15 & $2462.60/60.0$ & $214.48/4.0$ & $805.80/24.0$  & $1160.96/29.3$ & $99/103$ \\
    20 & $1159.36/28.0$ & $58.72/0.0$  & $643.60/36.0$  & $620.56/21.3$  & $103/103$ \\
    \bottomrule
  \end{tabular}
\end{table*}

Every evaluated suppression ratio lowers the macro-average generation length
and repetition rate relative to no defense, but the improvement is not
monotonic. At $\rho=20\%$, the macro-average output is shortest
(620.56 tokens), while $\rho=1\%$ and $\rho=20\%$ tie for the lowest macro-average
repetition rate (21.3\%). The attack-level results expose important
heterogeneity: RECITE is strongly mitigated at every evaluated ratio, GCG
benefits most at $\rho=20\%$.

\subsection{Effect of the Suppression Ratio on Benign Utility}
\label{app:ablation_suppression_ratio}

Table~\ref{tab:ablation_suppression_ratio} is a separate benign-utility study. Here, the learned direction ranking,
the training-set size, and all other training settings are fixed, and we vary
only the suppression ratio $\rho$, i.e., the fraction of hidden coordinates
included in the soft mask. 

\begin{table}[htbp]
  \centering
  \caption{Effect of the suppression ratio $\rho$ on benign task. Accuracy is reported in percent.}
  \label{tab:ablation_suppression_ratio}
  \small
  \setlength{\tabcolsep}{12pt}
  \begin{tabular}{lrr}
    \toprule
    $\rho$ (\%) & Accuracy & Avg. length \\
    \midrule
    0  & 82.0 & 2.125 \\
    1  & 82.0 & 2.130 \\
    5  & 81.5 & 2.145 \\
    10 & 82.0 & 2.115 \\
    15 & 81.0 & 2.125 \\
    20 & 80.5 & 2.065 \\
    \bottomrule
  \end{tabular}
\end{table}

Across the evaluated range, task accuracy decreases gradually as the suppression ratio increases, with a maximum drop of only 1.5 percentage points relative to the unsuppressed model. Meanwhile, the average response length varies by no more than 0.1 tokens.

\subsection{Sensitivity to the Maximum Repetition Distance}
\label{app:ablation_repetition_distance}

Finally, we vary the maximum token distance $q_{\max}$ considered by the
repeated-$n$-gram detector. This control determines how far apart matching
positions may be when constructing the tokenwise comparisons used by TRC.

\begin{table}[htbp]
  \centering
  \caption{Top-1 start layer under different maximum repetition distances
  $q_{\max}$. Layers are zero-indexed.}
  \label{tab:ablation_repetition_distance}
  \small
  \setlength{\tabcolsep}{9pt}
  \begin{tabular}{lccccc}
    \toprule
    Attack & $q_{\max}=1$ & $q_{\max}=2$ & $q_{\max}=3$ & $q_{\max}=4$ & $q_{\max}=5$ \\
    \midrule
    GCG    & 12 & 1 & 1 & 1 & 1 \\
    RECITE & 1  & 1 & 1 & 1 & 1 \\
    \bottomrule
  \end{tabular}
\end{table}

RECITE remains localized at layer 1 throughout, while GCG shifts from layer 12
at $q_{\max}=1$ to layer 1 for every $q_{\max}\geq2$. A one-token limit can be
too restrictive because a repeated semantic unit may be segmented into several
subword tokens; comparisons restricted to adjacent single-token matches can
therefore miss the actual cycle alignment. Allowing even a short multi-token
distance is sufficient to recover the stable shallow location in this
experiment. This result motivates using a small but non-unit search range
rather than assuming that one semantic repetition always corresponds to one
token.

\section{Online Efficiency}
\label{app:online_efficiency}

TRC does not append network layers, auxiliary detectors, or additional model forward passes at deployment. After offline localization, the fixed intervention is applied directly to the residual stream at the selected target layers by suppressing the identified residual coordinates during the original residual update. All other layers and residual coordinates remain unchanged. TRC therefore operates within the model's existing residual pathway, without introducing a separate inference module, additional forward computation, or changes to the depth of the computational graph.

Table~\ref{tab:llm_attack_time} reports attack-generation time for two LLMs.
TRC shortens generation in every listed model--attack condition. The reductions
range from 90.9\% to 99.6\%, because suppressing uncontrolled repetition allows
generation to terminate much earlier instead of continuing the attack-induced
loop.

\begin{table}[htbp]
  \centering
  \caption{Attack-generation time on two LLMs. Reduction is computed as
  $(T_{\mathrm{orig}}-T_{\mathrm{TRC}})/T_{\mathrm{orig}}$. Lower is better.}
  \label{tab:llm_attack_time}
  \small
  \setlength{\tabcolsep}{7pt}
  \begin{tabular}{llrrr}
    \toprule
    Model & Attack & Original (s) & TRC (s) & Reduction \\
    \midrule
    Qwen2.5-3B & Direct  & 278.92 & 9.33 & 96.7\% \\
    Qwen2.5-3B & GCG     & 402.52 & 1.73 & 99.6\% \\
    Qwen2.5-3B & LoopLLM & 91.72  & 8.36 & 90.9\% \\
    Llama-3.2-3B & Direct  & 340.19 & 5.38 & 98.4\% \\
    Llama-3.2-3B & GCG     & 64.73  & 1.67 & 97.4\% \\
    Llama-3.2-3B & LoopLLM & 143.44 & 4.68 & 96.7\% \\
    \bottomrule
  \end{tabular}
\end{table}

Table~\ref{tab:llm_mean_throughput} separately summarizes mean throughput. Across the two LLMs, throughput changes only slightly after intervention, decreasing by 3.00 tokens/s for Qwen2.5-3B and increasing by 1.05 tokens/s for Llama-3.2-3B. These small variations indicate that the intervention has negligible impact on generation throughput.

\begin{table}[htbp]
  \centering
  \caption{Mean throughput on two LLMs.
  Difference is TRC minus Original. Throughput is measured in tokens per second.}
  \label{tab:llm_mean_throughput}
  \small
  \setlength{\tabcolsep}{10pt}
  \begin{tabular}{lrrr}
    \toprule
    Model & Original & TRC & Difference \\
    \midrule
    Qwen2.5-3B   & 22.60 & 19.60 & $-3.00$ \\
    Llama-3.2-3B & 25.76 & 26.80 & $+1.05$ \\
    \bottomrule
  \end{tabular}
\end{table}

\end{document}